\documentclass[10pt,conference]{IEEEtran}

\usepackage{cite}
\ifCLASSINFOpdf
   \usepackage[pdftex]{graphicx}
   \usepackage{adjustbox}
   \graphicspath{{figs/}}
   \DeclareGraphicsExtensions{.pdf,.jpeg,.png}
\else
   \usepackage[dvips]{graphicx}
   \graphicspath{{../figs/}}
   \DeclareGraphicsExtensions{.eps}
\fi
\usepackage[cmex10]{amsmath}
\usepackage{amsthm}

\usepackage{algorithmic}
\usepackage{hyperref}
\usepackage{array}
\ifCLASSOPTIONcompsoc
  \usepackage[caption=false,font=normalsize,labelfont=sf,textfont=sf]{subfig}
\else
  \usepackage[caption=false,font=footnotesize]{subfig}
\fi
\usepackage{url}
\newif\iffinal
\finaltrue
\newcommand{\cmtid}{99999}
\iffinal
\else
\usepackage[switch]{lineno}
\fi

\begin{document}
%
% Copyright Notice
\thispagestyle{empty}
\onecolumn
\linespread{1.2}\selectfont{}
{\noindent\Huge IEEE Copyright Notice}\\[1pt]

{\noindent\large Copyright (c) 2024 IEEE

\noindent Personal use of this material is permitted. Permission from IEEE must be obtained for all other uses, in any current or future media, including reprinting/republishing this material for advertising or promotional purposes, creating new collective works, for resale or redistribution to servers or lists, or reuse of any copyrighted component of this work in other works.}\\[1em]

{\noindent\Large Accepted to be published in: IEEE Geoscience and Remote Sensing Letters, 2024.}\\[1in]

{\noindent\large Cite as:}\\[1pt]

{\setlength{\fboxrule}{1pt}
 \fbox{\parbox{0.65\textwidth}{P. E. C. Silva and J. Almeida, ``An Edge Computing-Based Solution for Real-Time Leaf Disease Classification using Thermal Imaging''. \emph{IEEE Geoscience and Remote Sensing Letters}, pp. 1--5, 2024, doi: 10.1109/LGRS.2024.3456637}}}\\[1in]
 
{\noindent\large BibTeX:}\\[1pt]

{\setlength{\fboxrule}{1pt}
 \fbox{\parbox{0.95\textwidth}{
 @Article\{GRSL\_2024\_Silva,
 
 \begin{tabular}{lll}
  & author    & = \{P. E. C. \{Silva\} and 
                    J. \{Almeida\}\},\\
			   
  & title     & = \{An Edge Computing-Based Solution for Real-Time Leaf Disease Classification \\
  &           & \ \ \ \ using Thermal Imaging\},\\
			   
  & journal   & = \{\{IEEE\} Geoscience and Remote Sensing Letters\},\\
  
  % & volume    & = \{\},\\
  
  % & number    & = \{\},\\
  
  & pages     & = \{1--5\},\\
  
  & year      & = \{2024\},\\
  
  & publisher & = \{\{IEEE\}\},\\
  
  & doi       & = \{10.1109/LGRS.2024.3456637\},\\
  \end{tabular}
  
\}
 }}}

\twocolumn
\linespread{1}\selectfont{}
\clearpage

\title{An Edge Computing-Based Solution for Real-Time Leaf Disease Classification using Thermal Imaging}


\iffinal

    \author{\IEEEauthorblockN{Públio Elon Correa da Silva}
    \IEEEauthorblockA{Federal University of São Carlos - UFSCar\\
    Email: \small\tt{publio@estudante.ufscar.br}}
    \and
    \IEEEauthorblockN{Jurandy Almeida}
    \IEEEauthorblockA{Federal University of São Carlos - UFSCar\\
    Email: \small\tt{jurandy.almeida@ufscar.br}}
    }


%Telephone: (800) 555--1212, Fax: (888) 555--1212}

\else
  \author{Sibgrapi paper ID: \cmtid \\ }
  \linenumbers
\fi

\maketitle

\begin{abstract}
Deep learning (DL) technologies can transform agriculture by improving crop health monitoring and management, thus improving food safety. In this paper, we explore the potential of edge computing for real-time classification of leaf diseases using thermal imaging. We present a thermal image dataset for plant disease classification and evaluate deep learning models, including InceptionV3, MobileNetV1, MobileNetV2, and VGG-16, on resource-constrained devices like the Raspberry Pi 4B. Using pruning and quantization-aware training, these models achieve inference times up to 1.48x faster on Edge TPU Max for VGG16, and up to 2.13x faster with precision reduction on Intel NCS2 for MobileNetV1, compared to high-end GPUs like the RTX 3090, while maintaining state-of-the-art accuracy.
\end{abstract}


\IEEEpeerreviewmaketitle

\section{Introduction}
\label{sec:intro}

Diseases in plants caused by living entities can be classified as biotic and abiotic~\cite{Kaur2019}. The primary sources of these biotic diseases are fungi, bacteria, and viruses. Conversely, abiotic diseases arise from non-living environmental factors, including hail, spring frosts, various weather conditions, exposure to chemicals, and more. These abiotic diseases, being non-infectious and non-transmissible, are generally less harmful and often preventable. 

Plant diseases pose a significant threat to food security, requiring prompt detection. However, the complexity of these diseases often challenges even experienced agronomists and plant biologists, leading to potential misdiagnosis and improper treatment~\cite{Batchuluun2022a}. In São Paulo, Brazil, crops are particularly vulnerable to climate variability, with extreme weather events like droughts, heat, and frost further complicating agricultural productivity. To address these challenges, efficient agricultural techniques such as crop rotation, cover cropping, and water management are essential for safeguarding and enhancing productivity~\cite{faesp2024},\cite{Nascimento2023}.

In addition, the significant reduction in crop yield is due to the inability to detect diseases early on, which invariably leads to decreased agricultural output. Consequently, the early recognition and examination of crop diseases are essential for maintaining crop quality. With recent advancements in computing speed and power, the use of large datasets has enhanced the efficiency of these systems~\cite{Paymode2022}. Therefore, the development of an automated process for leaf disease detection can also help with the reduction of agrochemicals and pesticide.

Most research on leaf disease classification has focused on visible images~\cite{Khan2022,GonzalezVictor2020},\cite{Shahidur2021,Brahimi2017}, which are sensitive to light conditions and may underperform without daylight. In contrast, thermal imaging, with its potential for overcoming these limitations, has been widely used in studies on environmental stress, crop yield, and seed vigor~\cite{Wen2023}.

Moreover, the use of traditional cloud infrastructures approaches for image classification poses a latency challenge between an edge device and the cloud~\cite{Varghese2016}, which is not feasible for real-time image classification. In this context, edge computing is a paradigm that facilitates computation near data sources, allowing processing at the edge layer instead of in the cloud. This approach is particularly advantageous when handling numerous distributed data sources where cloud computing may not be feasible~\cite{Zhang2019}.

Therefore, in this study we leverage the use of an embedded device such as the Raspberry Pi 4B as an edge device and perform aggressive model compression by means of pruning-quantization aware  training (PQAT) in addition of hardware acceleration devices such as the Edge TPU and Intel NCS2 to perform real-time image classification of leaf diseases using thermal camera.

Our main contributions can be summarized as follows:
\begin{itemize}
    \item \textbf{Introduction of a new dataset:} We introduce a thermal image dataset with leaf diseases across 16 types of plants.
    \item \textbf{Evaluation of model optimization techniques:} We evaluate pruning and quantization-aware training on MobileNetV1, MobileNetV2, VGG16, and InceptionV3, optimizing these models for performance on resource-constrained devices, such as the Raspberry Pi 4B.
    \item \textbf{Real-time classification hardware-based solution:} We develop a hardware-based solution that enables real-time image classification using hardware acceleration devices such as the Edge TPU (Coral USB) and the Visual Processing Unit (VPU), specifically the Intel NCS2.
\end{itemize}

\section{Related Work}
The work in~\cite{Atila2021} compared the performance of EfficientNet B5 and B4, with AlexNet and ResNet50, in classifying plant leaf diseases using transfer learning on the PlantVillage dataset. EfficientNet models achieved over 99\% accuracy in leaf disease detection. In \cite{Singh2022}, a 19-layer convolutional neural network~(CNN) effectively detected Marsonina Coronaria and Apple Scab with 50,000 images. This model outperformed other CNNs and machine learning classifiers, achieving 99.2\% accuracy, underscoring the effectiveness of deep learning~(DL) in agricultural disease classification.

A study in~\cite{Batchuluun2022a} presented an approach for plant classification. It leveraged DL methods and used a dataset comprising nonaligned thermal and visible light images of plants. This study introduced a new DL model, called PlantCR, which combines features extracted from both thermal and visible light images to enhance classification accuracy. The proposed model achieved better multi-class classification performance compared to previous methods that relied solely on thermal or visible light images.

The idea of using thermal images for leaf diseases detection is further expanded in~\cite{Batchuluun2023}, which focused on using thermal images for plant and crop disease classification. It proposed a novel method using a CNN with explainable artificial intelligence (XAI) to enhance accuracy in these classifications. This study introduced a new thermal plant image dataset and also utilized an open database of crop diseases. The proposed method achieved high classification accuracy on both datasets, outperforming existing methods.

Few studies have explored thermal images for leaf disease classification, and only one publicly available dataset exists for thermal paddy leaves, with 636 samples and 6 classes. To address this scarcity and support further research, we introduce a new publicly available dataset with 15,144 images across 7 classes, featuring leaf diseases from 16 plant species common in São Paulo, Brazil.

\section{Types of Leaf Disease}
\begin{figure}[!htb]
  \centering
  \includegraphics[width=0.75\linewidth]{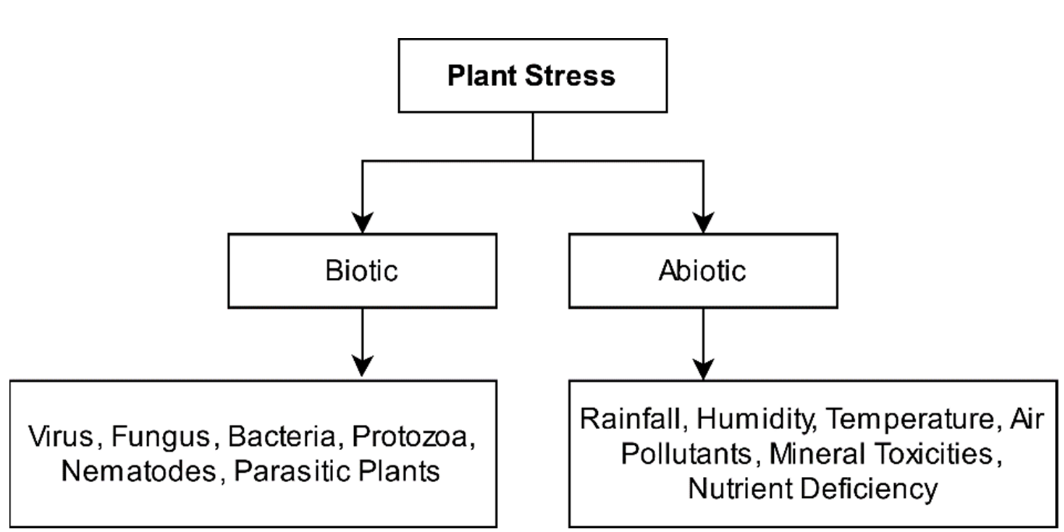} 
  \caption{Types of plant stress}
  \label{fig:stress_types}
\end{figure}

Diseased plants show symptoms like color changes, altered shape and size, and growth retardation, varying across disease stages. During the transition stage, disease factors begin affecting healthy plants, making it difficult to distinguish between healthy and diseased plants. Diseases can weaken plant immune systems, increasing susceptibility to multiple diseases and leading to similar symptoms across different diseases, as shown in Fig.~\ref{fig:stress_types}.

In leaf blight diseases, leaf tips and edges first yellow—a process known as chlorosis—before necrosis sets in, creating dark-brown patches that eventually lead to complete leaf death. Older leaves are typically more susceptible~\cite{Li2019}.

Leaf spot diseases significantly impact crop production and cause economic losses. Early detection is crucial for mitigation. Spots typically appear a week after onset as white to grayish-white with colored margins and can cause perforations in the leaf lamina. Newer leaves usually show the first signs of fungal lesions~\cite{Jain2019}.

Mosaic Virus in cassava and sugarcane causes significant crop damage and yield losses, spreading through whiteflies or vegetative propagation. Symptoms like chlorotic patterns, leaf distortion, and size reduction can be difficult to identify due to environmental factors. Although advanced AI methods like CNNs improve disease detection~\cite{Dhaka2021}, the presence of irrelevant information in images, such as dead leaves, soil, or weeds, remains a challenge for DL techniques.

\section{Edge Computing}
\begin{figure}[!htb]
  \centering
  \includegraphics[width=1\linewidth]{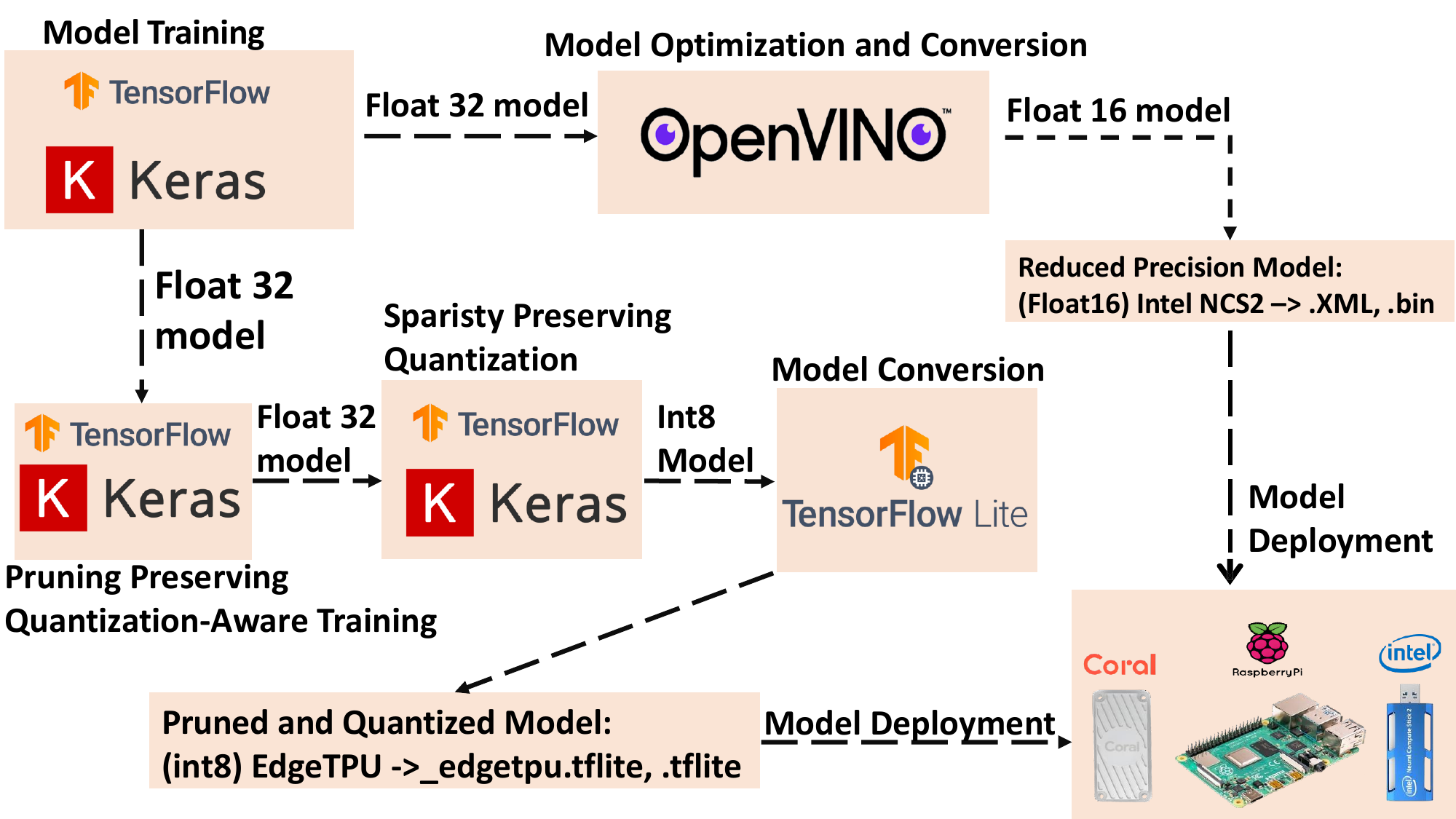}
  \caption{General workflow for deploying models to the Raspberry Pi at the edge.}
  \label{fig:edge_deploy}
\end{figure}

Edge Computing~(EC) brings data processing closer to the source, making it ideal for real-time applications where inference speed is critical. By processing data on local devices like IoT devices or edge servers, this approach reduces latency and bandwidth demands. With advanced computation platforms, DL is now widely applied in AI tasks, like image classification and object recognition~\cite{Abirami2020}.
In this context, EC allows farmers to use drone-mounted thermal cameras to detect leaf diseases in real time by classifying frames on-site. This is essential, as cloud-based approaches can introduce latency due to limited bandwidth or connectivity issues, especially in dead zones. Such delays risk unclassified frames and missed detections. Local processing enables prompt classification, reducing the risk of missed frames and supporting timely agricultural interventions.
Additionally, model compression allows resource-constrained devices to gain inference speed with reduced bandwidth consumption while trading off a very small accuracy due to the model precision reduction, which can be considerably advantageous especially for tasks like real-time image classification~\cite{Kristiani2020}. 
Fig.~\ref{fig:edge_deploy} shows the general workflow used to train and deploy models to a Raspberry Pi for model inference with the help of accelerators.

In this workflow depicted above, model training starts with Float32 precision, followed by precision reduction to int8 for Coral TPU and Float16 for Intel NCS2. This step optimizes computational efficiency and model performance. Notably, the accelerators, such as the Coral TPU and Intel NCS2, are designed exclusively for accelerating inference and do not support model training. Furthermore, training directly on the Raspberry Pi is not feasible for fast model training due to its limited computational power. After optimization, models are accelerated and formatted for deployment: TFLite files for Coral TPU and OpenVINO's XML and bin files for the NCS2.

\section{Real-time Image Classification}
Real-time image processing relies on minimizing computational demands and enhancing algorithm efficiency, considering hardware constraints~\cite{Moghimi2021}. Given that edge devices often have limited resources, various optimization techniques and on-device acceleration methods have been explored for DL algorithms~\cite{Swarnava2019}.

Pruning improves neural network efficiency by removing low-impact weights, reducing model size and computational needs without significantly affecting accuracy. Similarly, quantization techniques convert neural network precision from floating points to low-bit integers, optimizing models for resource-constrained devices.

Training-aware quantization preserves accuracy by integrating quantization during training, while post-training quantization applies methods like weight sharing and pruning. Fine-tuning with small datasets can further improve accuracy~\cite{ChoudharyT2020, LiangT2021}. Both techniques reduce model size and complexity, facilitating deployment on devices with limited capacity.

\section{Materials and Methods}
This section details the transfer learning method for real-time leaf disease classification using thermal imaging, employing MobileNetV1, MobileNetV2, VGG-16, and InceptionV3 models pretrained with ImageNet weights. These models were selected for their compatibility with the Edge TPU and Intel NCS2. The dataset was split into 60\% for training, 20\% for validation, and 20\% for testing, and the models were trained separately before deployment to the Raspberry Pi.

\begin{figure}[!htb]
  \centering
  \includegraphics[width=0.85\linewidth]{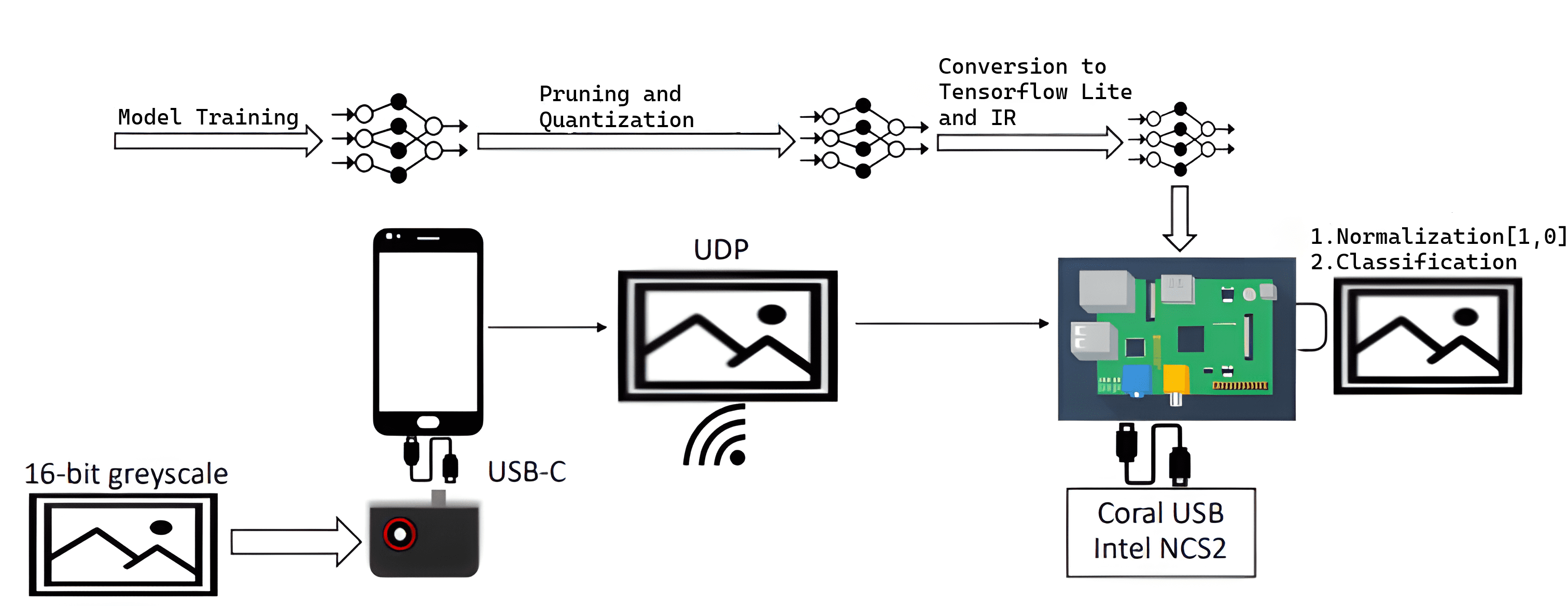}
  \caption{The pipeline of the proposed hardware based solution for leaf disease classification.}
  \label{fig:proposed}
\end{figure}

Fig.~\ref{fig:proposed} illustrates the proposed hardware based solution for leaf disease classification.
A mobile phone equipped with a thermal camera connected via USB is used. The thermal camera uses an android application with a real-time streaming feature to capture greyscale thermal images of the plant's leaf. This image is then streamed by the mobile app to the Raspberry Pi 4B, where it is normalized, resized, and used as input for a classifier. The classifier, accelerated by a USB device connected to the edge server, processes the image in real-time to obtain labels and confidence levels.

\subsection{Dataset Collection}
\begin{figure}[!htb]
  \centering
\includegraphics[width=0.82\linewidth]{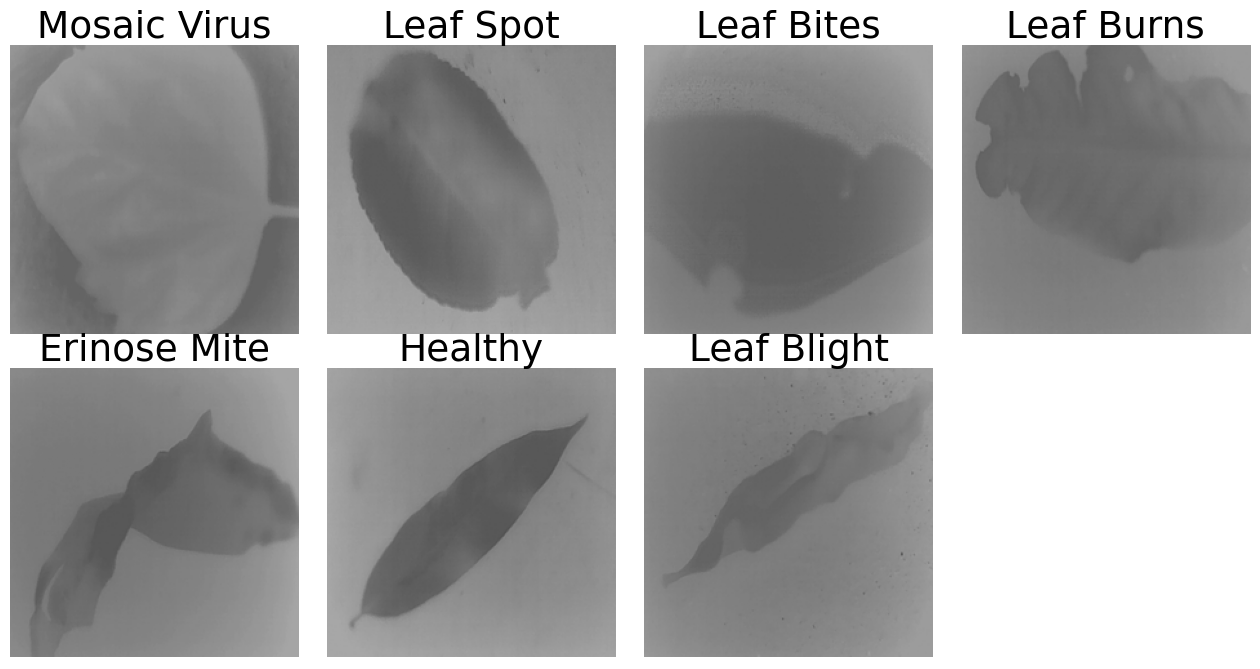}
  \caption{Image samples from the self collected dataset.}
  \label{fig:dataset}
\end{figure}

To the best of our knowledge, the only publicly available dataset currently is the thermal paddy leaves dataset~\cite{Batchuluun2022a}. Motivated by this, we propose a new thermal dataset of diseases containing 15,444 images and 7 classes. Fig.~\ref{fig:dataset} shows samples from each class. Every image representing different leaf diseases from each plant species were grouped by their common leaf disease. 

The collected dataset includes species like Persea americana, Malpighia emarginata, Myrciaria glazioviana, among others, such as Zantedeschia aethiopica and Litchi chinensis, extending to Citrus limon and Mangifera indica. The code and datasets used in this study are available in our \href{https://github.com/publioelon/Leaf-Diseases-Classification}{GitHub}.

The thermal imaging dataset was captured using an Infiray T3C thermal camera, which records frames in grayscale and operates in a spectral range of 8-14µm. This camera features a resolution of 384x288 pixels, a frame rate of 25Hz, manual focus adjustment, and a measurement accuracy of ±2\%.

\subsection{Experimental Protocol}
A NVIDIA RTX3090 GPU with an Intel Core i5 11400 processor and 64GB RAM was used for training, and inference was conducted on a Raspberry Pi 4B with 8GB RAM. All models used Adam and categorical cross-entropy loss for a fair comparison. The hyperparameters yielding the highest validation accuracy are listed in Table~\ref{tab:basic-hyperparameters}.

\begin{table}[!htb]
\centering
\caption{Training Hyperparameters.}
\label{tab:basic-hyperparameters}
\resizebox{\columnwidth}{!}{%
\begin{tabular}{|l|c|c|c|c|}
\hline
\textbf{Param.} & \textbf{MobileNet} & \textbf{MobileNetV2} & \textbf{VGG16} & \textbf{IncepV3} \\ \hline
Batch Size      & 32                 & 128                  & 32             & 32               \\ \hline
LR              & 1e-4               & 5e-6                 & 5e-5           & 1e-6             \\ \hline
Epochs          & 5                  & 120                  & 25             & 150              \\ \hline
Dropout         & 0.5                & 0.5                  & 0.4            & 0.5              \\ \hline
\end{tabular}
}
\end{table}

To mitigate overfitting, data augmentation was implemented with the following operations: variable zoom ranges with values between 10\% and 20\%, horizontal and vertical shifts, image flipping, rotation ranging from 20 to 90 degrees, and brightness adjustment with values between 0.9 and 1.3. 
In addition, fully connected layers were also added in the MobileNetV1, MobileNetV2, and InceptionV3 models, with dropout rates ranging from 128 to 256.
Table~\ref{tab:combined-hyperparams} shows the hyperparameters used during the pruning and quantization steps to compress the models for the Edge TPU device before the TensorFlow Lite conversion.

\begin{table}[!htb]
\centering
\caption{PQAT Fine-Tuning Hyperparameters for VGG16, InceptionV3, MobileNetV1, and MobileNetV2.}
\label{tab:combined-hyperparams}
\begin{tabular}{|l|c|c|c|c|}
\hline
\textbf{Param.} & \textbf{VGG16} & \textbf{IncepV3} & \textbf{MobileNetV1} & \textbf{MobileNetV2} \\
\hline
PLR & 5e-6 & 1e-5 & 1e-5 & 5e-6 \\
\hline
Spar. & 0.7 & 0.5 & 0.7 & 0.1 \\
\hline
Epochs & 1 & 20 & 1 & 5 \\
\hline
QLR & 5e-6 & 1e-5 & 1e-6 & 5e-7 \\
\hline
Q. Epochs & 1 & 5 & 5 & 5 \\
\hline
\end{tabular}
\\
\footnotesize{Abbreviations: LR - Learning Rate. PLR - Pruning Learning Rate, Spar. - Target Sparsity, Epochs - Pruning Epochs, QLR - Quantization Learning Rate, Q. Epochs - Quantization Epochs.}
\end{table}

Experiments were conducted with TensorFlow, TensorFlow Lite, and Edge TPU at standard and maximum frequencies to evaluate model performance using the following procedures:
\begin{enumerate}
    \item \textbf{Inference Time:} We calculated inference times for all 3340 test samples, obtaining mean and standard deviation values, with the first inference time measured separately due to its typically longer duration.
    \item \textbf{Frames per Second:} To calculate average FPS, only the time for receiving and classifying images at the edge server was considered, using 3340 frames sent from a stationary thermal camera focused on a wall.
    \item \textbf{Test Accuracy:} Accuracy was determined by the sum of correct predictions within the test set.
\end{enumerate}

\section{Results}
\begin{figure}[!htb]
    \centering
    \adjustbox{trim=0mm 3mm 0mm 5mm}{%
        \includegraphics[width=\columnwidth]{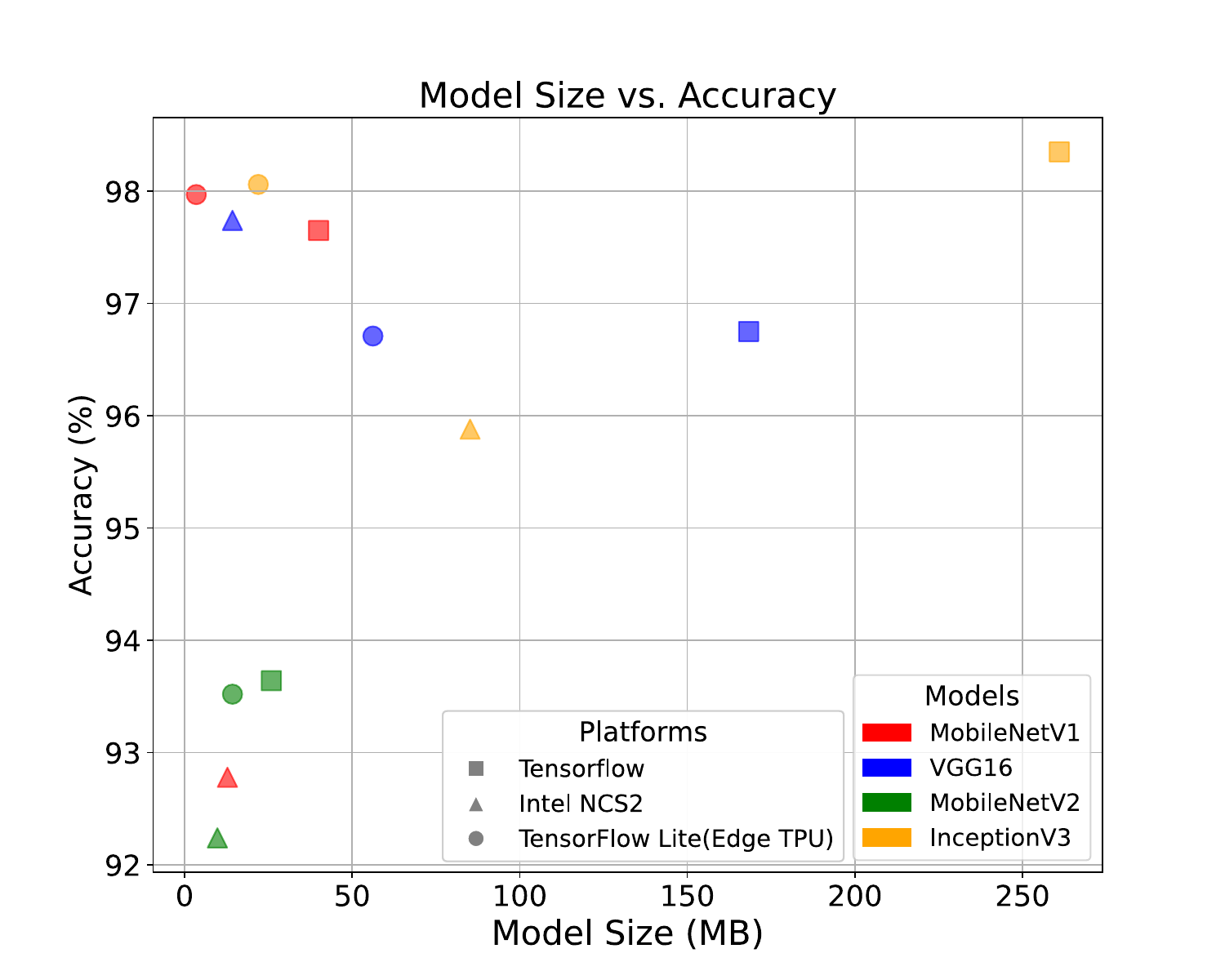}
    }
    \caption{Model Size vs. Accuracy for Different Platforms}
    \label{fig:model_size_vs_accuracy}
\end{figure}

In this work, four DL architectures for real-time leaf disease classification utilizing thermal imaging were evaluated, namely: MobileNet, MobileNetV2, InceptionV3, and VGG-16. The experimental results demonstrate an overall increase in inference speed for all the models. Notably, the models that deliver the best performance in terms of accuracy and size compression are MobileNetV1 and InceptionV3, both pruned and quantized to int8 for the Coral USB accelerator, as well as VGG-16, which is optimized with reduced precision to float16 for OpenVino, as denoted in Fig.~\ref{fig:model_size_vs_accuracy}. 

\begin{figure}[!htb]
  \centering
  \includegraphics[width=\columnwidth]{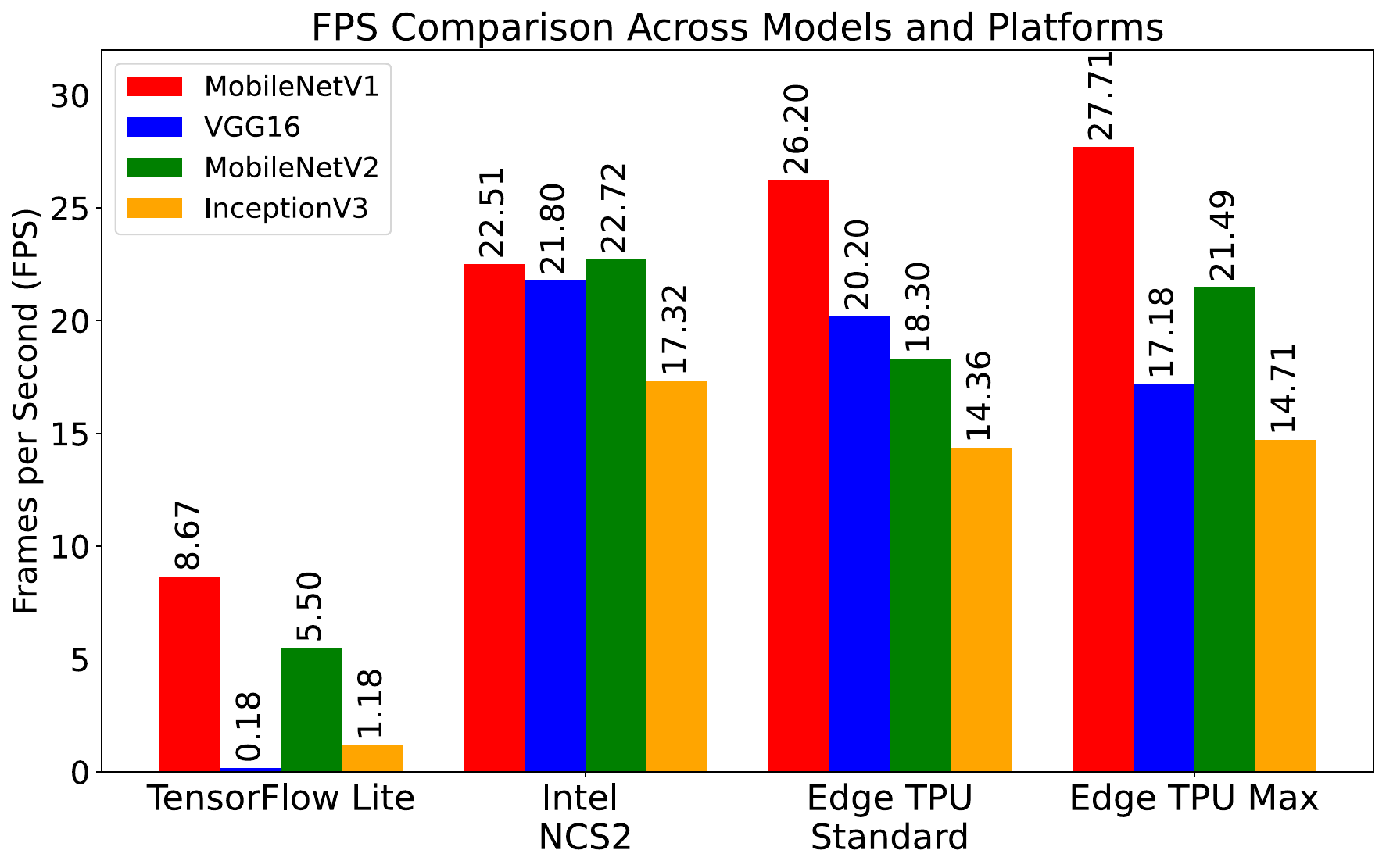}
  \caption{Frames per second.}
  \label{fig:fps}
\end{figure}

Fine-tuning after pruning-quantization slightly improved the accuracy of MobileNetV1 and MobileNetV2 by removing redundant parameters~\cite{LiangT2021}. InceptionV3, which saw the largest size reduction, also benefited from PQAT. As shown in Fig.~\ref{fig:fps}, MobileNetV1 is well-optimized for resource-constrained devices, achieving 8 FPS even without acceleration. The Coral USB TPU further enhanced the Raspberry Pi 4B's performance, allowing it to process a higher frame rate than the thermal camera with MobileNetV1. MobileNetV2 saw a greater frame rate increase with Edge TPU at max frequency compared to TensorFlow Lite, and PQAT's aggressive compression also significantly improved frame rates for InceptionV3 and VGG-16.

\begin{figure}[!htb]
  \centering
  \includegraphics[width=\columnwidth]{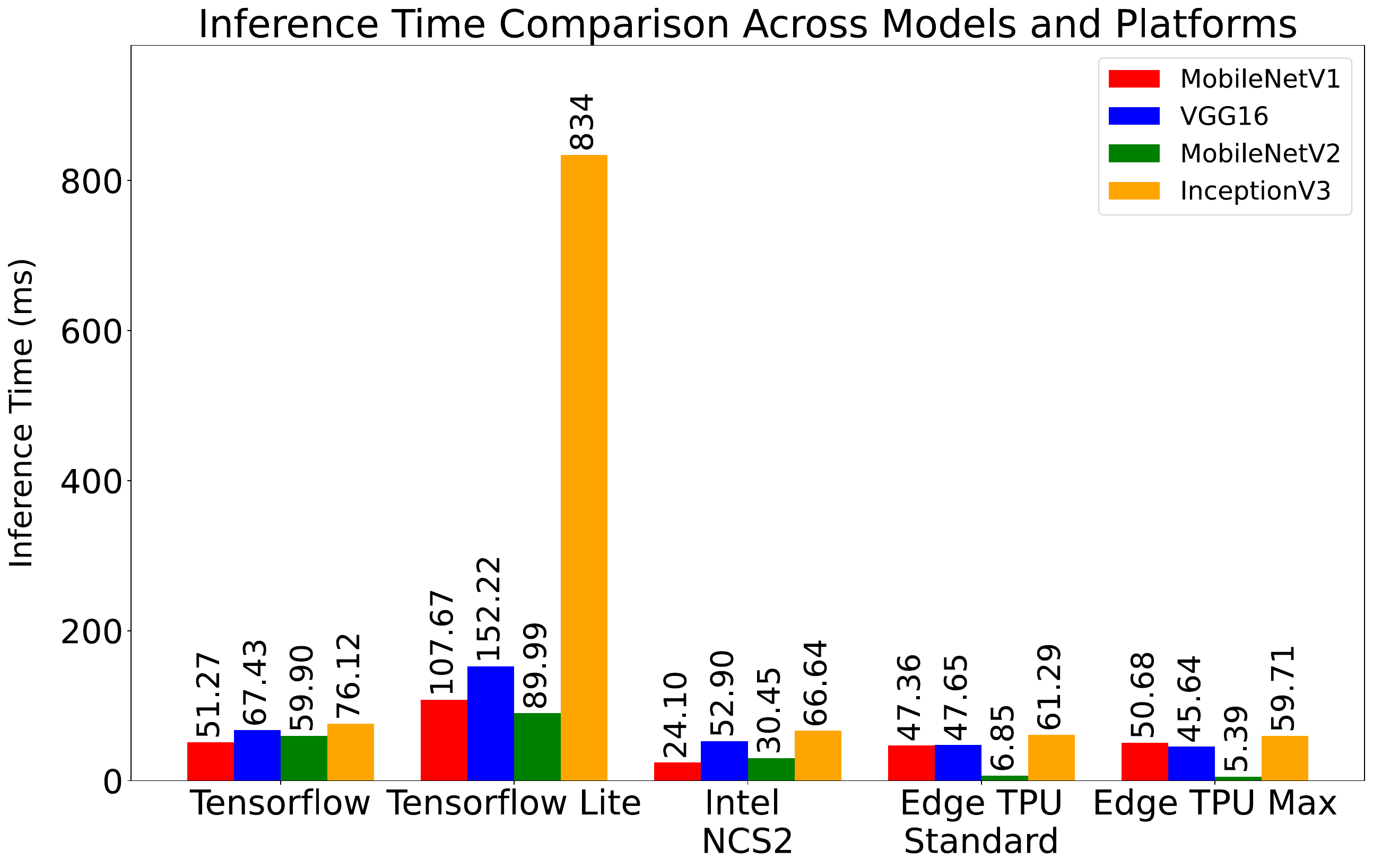}
  \caption{Inference Time.}
  \label{fig:inference}
\end{figure}

Interestingly, it can be seen in Fig.~\ref{fig:inference} that the compressed models yielded a faster inference speed than those tested on Tensorflow, which were run on a RTX 3090 GPU. Also, the InceptionV3 model had a considerable reduction in the inference time when using the accelerators, which denotes the feasibility of using a complex model such as InceptionV3 on a resource-constrained device like the Raspberry Pi 4B.

The inference times for MobileNetV1 and MobileNetV2 on the Raspberry Pi 4B with Edge TPU (5.9 ms and 7.0 ms, respectively) align closely with the results in~\cite{Jainuddin2020}, even with the addition of fully connected layers in our models. Notably, the referenced study did not explore increasing the Edge TPU's frequency, which could enhance performance further. Regarding model compression, InceptionV3 and VGG-16 achieve similar compression sizes as reported in~\cite{Rosero-Montalvo2024}. However, model sizes can vary based on the training dataset, a point highlighted by the authors. Additionally, their analysis does not account for OpenVino's optimization tool, which reduces model precision to Float16, making models more compact for deployment on edge devices.

In summary, our results demonstrated the application of PQAT on four state-of-the-art CNNs. 
Our goal was to optimize these models for deployment on a Raspberry Pi. 
Due to resource constraints, a balance between model complexity and predictive performance is required.
Overall, PQAT achieved promising results in preparing such models for deployment on resource-constrained embedded devices, like Raspberry Pi 4B.

\section{Conclusion}
This paper demonstrates the use of edge computing for real-time leaf disease classification with thermal imagery, introducing a novel dataset. Our approach surpasses high-end GPUs like the RTX 3090 in efficiency, though thermal cameras' sensitivity to daylight heat may affect accuracy. The method, currently limited to detecting a single disease per leaf, was optimized through pruning and quantization-aware training, maintaining or improving accuracy. This study highlights the effectiveness of deploying compressed models on devices like the Raspberry Pi 4B with hardware acceleration, reducing reliance on cloud computing. Future work could explore image fusion with thermal and visible spectrum images, evaluate additional edge devices like the Jetson Nano for comparison with the Edge TPU and NCS2, and investigate multi-label object detection to address current limitations.


\section*{Acknowledgment}
This research was supported by São Paulo Research Foundation - FAPESP (grant 2023/17577-0), Brazilian National Council for Scientific and Technological Development - CNPq (grants 315220/2023-6 and 420442/2023-5), and Coordination for the Improvement of Higher Education Personnel (CAPES) – Funding Code 001.



% \bibliographystyle{IEEEtran}
% \bibliography{refs}

\end{document}